\documentclass{article}

\usepackage[accepted]{icml2023}
\usepackage[textsize=tiny]{todonotes}
\usepackage{bm}
\usepackage{graphicx}
\usepackage{subfigure}
\usepackage{url}

\usepackage{microtype}
\usepackage{graphicx}
\usepackage{caption}
\usepackage{booktabs} 
\usepackage{color}
\usepackage{tabularx}
\usepackage{hyperref}

\usepackage{amsmath}
\usepackage{amssymb}
\usepackage{mathtools}
\usepackage{amsthm}

\usepackage[capitalize,noabbrev]{cleveref}

\def\eqref#1{equation~\ref{#1}}

\def\1{\bm{1}}

\def\vx{{\bm{x}}}

\def\vz{{\bm{z}}}

\DeclareMathAlphabet{\mathsfit}{\encodingdefault}{\sfdefault}{m}{sl}
\SetMathAlphabet{\mathsfit}{bold}{\encodingdefault}{\sfdefault}{bx}{n}

\def\gD{{\mathcal{D}}}

\def\gG{{\mathcal{G}}}

\newcommand{\E}{\mathbb{E}}
\newcommand{\Ls}{\mathcal{L}}

\theoremstyle{plain}

\theoremstyle{definition}

\theoremstyle{remark}

\newcommand{\myparagraph}[1]{
\vspace{0cm}\noindent
\textbf{#1.}
}

\newcommand{\SD}{\mathbb{SD}}

\newcommand{\lcal}{\mathcal{L}}

\newcommand{\ctgan}{\texttt{CTGAN}}
\newcommand{\tablegan}{\texttt{TableGAN}}
\newcommand{\tvae}{\texttt{TVAE}}
\newcommand{\mctgan}{\texttt{MargCTGAN}}

\newcommand{\cond}{\text{cond}}
\newcommand{\wass}{\text{WGP}}

\newcommand{\data}{\text{data}}
\newcommand{\mean}{\text{mean}}
\newcommand{\std}{\text{std}}

\newcommand{\Ereal}{\underset{\vx\sim p_{\data}}{\E}}
\newcommand{\Efake}{\underset{\vz\sim p_{\vz}}{\E}}

\renewcommand{\sectionautorefname}{Section}

\icmltitlerunning{MargCTGAN: A ``Marginally'' Better CTGAN for the Low Sample Regime}

\begin{document}
\twocolumn[
\icmltitle{MargCTGAN: A ``Marginally'' Better CTGAN for the Low Sample Regime}

\icmlsetsymbol{equal}{*}
\begin{icmlauthorlist}
\icmlauthor{Tejumade Afonja}{cispa}
\icmlauthor{Dingfan Chen}{cispa}
\icmlauthor{Mario Fritz}{cispa}
\end{icmlauthorlist}

\icmlaffiliation{cispa}{CISPA Helmholtz Center for Information Security, Saarbrücken, Germany}

\icmlcorrespondingauthor{Tejumade Afonja}{tejumade.afonja@cispa.de}
\icmlkeywords{Synthetic Data,  GAN, Tabular Data, Evaluation Metrics}
\vskip 0.3in
]
\printAffiliationsAndNotice{}  

\begin{abstract}
The potential of realistic and useful synthetic data is significant. However, current evaluation methods for synthetic tabular data generation predominantly focus on downstream task usefulness, often neglecting the importance of statistical properties. This oversight becomes particularly prominent in low sample scenarios, accompanied by a swift deterioration of these statistical measures. In this paper, we address this issue by conducting an evaluation of three state-of-the-art synthetic tabular data generators based on their marginal distribution, column-pair correlation, joint distribution and downstream task utility performance across high to low sample regimes. The popular $\ctgan$ model shows strong utility, but underperforms in low sample settings in terms of utility. To overcome this limitation, we propose $\mctgan$ that adds feature matching of de-correlated marginals, which results in a consistent improvement in downstream utility as well as statistical properties of the synthetic data.
\end{abstract}

\vspace{-6pt}
\section{Introduction}
\label{introduction}
Tabular data, despite being the most widely used data type~\cite{kagglesurvey}, presents substantial challenges ranging from data heterogeneity and quality measurement to imbalance and privacy concerns. Encouragingly, recent advancements in synthetic tabular data generators have shown considerable promise in tackling these issues. These models have shown effectiveness in handling heterogeneous data attributes~\cite{ctgan, ctab}, facilitating the safe sharing of personal records~\cite{medgan, tablegan,mottini2018airline}, and mitigating class imbalance~\cite{engelmann2021conditional}. Nonetheless, the evaluation of existing models predominantly focuses on downstream machine learning tasks and large datasets. This evaluation paradigm overlooks their utility in broader practical scenarios, especially the data-limited, low-resource settings, and fails to consider other crucial aspects of synthetic datasets including fidelity, diversity, and authenticity~\cite{alaa2022faithful}.

In response to these challenges, we introduce a comprehensive evaluation framework, integrating nine distinct metrics across four critical dimensions: downstream task utility, joint fidelity, preservation of attribute correlations, and alignment of marginals (\sectionautorefname~\ref{metric}). Our objective is to thoroughly evaluate the representative models using diverse metrics, aiming at a comprehensive understanding of their quality and adaptability, particularly for scenarios that are underexplored in existing literature.

Our evaluation uncovers intriguing insights into the characteristics of state-of-the-art models. For instance, the popular $\ctgan$~\cite{ctgan} model typically demonstrates high attribute fidelity but falls short in utility for low-data scenarios. Conversely, $\tablegan$~\cite{tablegan} exhibits better utility but lacks performance in other dimensions. To capitalize on the strengths of both models, we propose $\mctgan$ that improves upon $\ctgan$ by introducing feature matching of decorrelated marginals in the principal component space. This approach consistently improves utility without compromising other fidelity measures, especially in the data-limited settings.

\vspace{-6pt}
\section{Related Works}
\label{related_works}

\myparagraph{Tabular Data Generators}
In recent years, deep generative models have seen significant advancements in their application to diverse forms of tabular data, including discrete attributes~\cite{medgan}, continuous values~\cite{mottini2018airline}, and heterogeneous mixtures~\cite{tablegan,xu2018synthesizing,ctgan,ctab}. 
Notably, $\tablegan$~\cite{tablegan}, $\ctgan$~\cite{ctgan}, and $\tvae$~\cite{ctgan} stand out as state-of-the-art benchmark models and will be the focus of our empirical evaluation. 
On the other hand, the issue of limited data availability remains underexplored in literature, despite several attempts to bypass such challenges by effectively combining multiple data sources~\cite{chen2019faketables,ma2020vaem,yoon2018radialgan}. Our work aims to fill this research gap by introducing a systematic evaluation across various scenarios, ranging from full-resource to data-limited cases. Additionally, we propose model improvements that designed to effectively capture the underlying data structure in low-sample settings.

\myparagraph{Evaluation of Tabular Data Generators}
The evaluation of generators, particularly for tabular data, is a challenging area due to its requirement for complex metrics, unlike simpler visual inspection for image data~\cite{theis2016note}. Recent studies have introduced a variety of metrics: prominent among these are downstream machine learning efficacy, unified metrics evaluation~\cite{chundawat2022tabsyndex}, and evaluations focusing on distinct aspects~\cite{alaa2022faithful,dankar2022multi}. In this work, we consider comprehensive evaluation methods encompassing machine learning efficacy, statistical properties such as divergence on marginals, column correlations, and joint distance.

\vspace{-6pt}
\section{Method: $\mctgan$}
\label{methods}
$\mctgan$ adheres to the standard Generative Adversarial Networks (GANs) paradigm~\cite{goodfellowNIPS2014}, which involves training a generator $\gG$ and a discriminator $\gD$ in an adversarial manner. The training target is to enhance the discriminator's ability to distinguish between real and fake data, while simultaneously updating the generator to produce samples that are increasingly realistic. We adopt the WGAN-GP objective~\cite{gulrajani2017improved} in which the overall training process can be interpreted as optimizing the generator to minimize the Wasserstein distance between the distributions of the generated and real data:
\begin{equation*}
\resizebox{\columnwidth}{!}{
$\lcal_{\wass} = \Efake\big[\gD\big(\gG(\vz)\big)\big] -\Ereal \big[\gD(\vx)\big] + \lambda \big(\Vert \nabla_{\hat{\vx}}\gD(\hat{\vx})\Vert_2-1\big)^2$}
\end{equation*}
where $\hat{\vx}$ is constructed by interpolating real and generated samples and $\lambda$ denotes the weight for the gradient penalty. The discriminator is trained to minimize $\lcal_{\wass}$, while the generator is trained to maximize it.

Following the $\ctgan$~\cite{ctgan} framework, we adopt several key techniques to adapt GAN models for tabular data. Firstly, one-hot encoding is applied to pre-process categorical attributes, paired with the Gumbel-softmax function serving as the network output activation function, thereby ensuring differentiability. Secondly, for numerical attributes, we apply a technique known as \textit{mode-specific normalization} in the pre-processing phase, enabling an accurate reflection of the multi-modality in the values distribution. Lastly, we employ the \textit{training-by-sampling} strategy during the training process, which effectively balances the occurrences of different classes in the categorical columns to match their real distribution. This strategy introduces an additional loss term on the generator, which we denote as $\Ls_{\cond}$.

While $\ctgan$ generally demonstrates promising utility for training downstream machine learning classifiers, it often falls short in low-sample scenarios (See \figureautorefname~\ref{fig:averaged_metric_results}). Drawing inspiration from $\tablegan$
, we propose a moment matching loss that
proactively encourages the generator to learn and mirror the first and second-order data statistics. Notably, unlike $\tablegan$ which attempts to match statistics on the features extracted by the discriminator, we compute the first and second moments after conducting the Principal Component Analysis (PCA) on the data. 
Specifically, the transform is performed while maintaining the original data dimensionality, i.e., we simply decorrelate without down-projection. Intriguingly, this straightforward technique proves effective (\figureautorefname~\ref{fig:ablation}), likely because the decorrelated feature representation supports the independent moment matching.
Formally, 
\begin{align}
\mathcal{L}_{\mean} = &\Big\Vert \Ereal \big[f\left(\vx\right)\big] - \Efake
\big[f\big(\gG(\vz)\big)\big] \Big\Vert_2 \\
\mathcal{L}_\std = &\Big\Vert \underset{\vx\sim p_{\data}}{\SD}\big[f(\vx)\big] - \underset{\vz\sim p_\vz}{\SD}\big[f\big(\gG(\vz)\big)\big] \Big\Vert_2 \\
\lcal_{\text{marg}} = &\lcal_{\mean} + \lcal_\std \\
\lcal^{\gG} = &-\lcal_{\wass} + \lcal_{\cond} + \lcal_{\text{marg}} 
\end{align}
where $f(\cdot)$ denotes the PCA transformation function. $\Ls_{\mean}$ targets the mean, while $\Ls_{\std}$ focuses on the standard deviation. The total training losses for the generator and discriminator are  $\Ls^\gG$ and $\Ls_{\wass}$, respectively.

\vspace{-6pt}
\section{Multi-Dimensional Evaluation Metrics}
\label{metric}

We present a comprehensive evaluation that accesses tabular data generators performance across four critical dimensions: application utility, joint fidelity, column-pair fidelity, and marginal fidelity. The implementation details can be found in Appendix~\ref{implementation_details}.

\myparagraph{Application Utility}
This dimension focuses on the efficacy of synthetic data as a substitute for real data in specific tasks. This effectiveness is typically quantified by \textit{machine learning efficacy} that evaluates the performance (e.g., F1-score or accuracy) on a distinct real test dataset when training predictive ML models on synthetic data. In situations where knowledge of the target downstream task is unavailable, an alternate methodology known as \textit{dimension-wise prediction} (or \textit{all-models test}) may be employed. This methodology considers each column as a potential target variable for the task and reports the mean performance across all cases.

\myparagraph{Joint Fidelity}
This category aims to quantify the similarity between the overall \textit{joint} distributions of real and synthetic data. While an exact measurement is always intractable, the most commonly used approximation is the \textit{distance to closest record}. This computes the Euclidean distance between each synthetic data sample and its nearest neighbors in the real test dataset, intending to assess the possibility of each synthetic sample being real.
Conversely, the \textit{likelihood approximation} computes the distance between each real test sample and its closest synthetic sample. This mirrors the probability of each real sample being potentially generated by the model, thereby encapsulating a concept of data likelihood.

\myparagraph{Column-Pair Fidelity}
This dimension investigates the preservation of feature interactions, specifically focusing on the direction and strength of correlations between pairs of columns in the synthetic dataset as compared to the real dataset. A commonly used metric for this purpose is the \textit{association difference}, also referred to as the \textit{pairwise correlation difference}. This measure quantifies the discrepancy between the correlation matrices of the real and synthetic datasets, where the correlation matrix encapsulates the pairwise correlation of columns within the data.

\myparagraph{Marginal Fidelity}
A key prerequisite for accurately replicating the real data distribution is to ensure a match in the distribution of each individual column, i.e., aligning the marginals. The evaluation of this criterion essentially involves quantifying the disparity between two one-dimensional variables. Commonly used metrics for this purpose include the \textit{Jensen-Shannon divergence}, \textit{Wasserstein distance}, \textit{column correlation}, and \textit{histogram intersection}. 
While the divergence measures are directly computed for categorical attributes, numerical columns typically undergo a pre-processing discretization step via binning prior to divergence computation to ensure tractability.

\vspace{-6pt}
\section{Experiments}
\label{sec:experiments}

\myparagraph{Setup}
We conducted evaluations on four benchmark tabular datasets: Adult~\cite{adult}, Census~\cite{census},  News~\cite{news}, and Texas. 
These datasets exhibit diverse properties in terms of size (spanning $30$-$110$ thousand samples), column heterogeneity, and distinct characteristics (Refer to \tableautorefname~\ref{datasets} and Appendix~\ref{dataset_info} for details). Our investigation spans a geometric progression of sample sizes, extending from $40$ to the full dataset size (notated as ``$-1$''), to emulate a range from low to high resource settings. In line with existing studies~\cite{ctgan}, models were trained for $300$ epochs. The evaluations were conducted on a separate test set that was never used during the whole training process of the tabular data generators. To account for potential randomness, experiments were conducted over three different random seeds for model training and repeated across five trials for generating synthetic datasets. See details in Appendix~\ref{implementation_details}.

\begin{table}[!t]
\caption{
Summary of Datasets. \textbf{Col} refers to number of columns. \textbf{N/B/M} correspond to the number of numerical, binary, and multi-class categorical columns, respectively. 
}
\label{datasets}
\begin{center}
\begin{small}
\vspace{-10pt}
\begin{tabular}{lcccc}
\toprule
Dataset & Train/Test size & Col & N/B/M &Task \\
\midrule
Adult    & 34118/14622 & 15 & 7/2/6 & classification \\
Census & 199523/99762 & 41 & 7/3/31 & classification \\
News    & 31644/8000 & 60 & 45/15/0 & classification \\
Texas    & 60127/15032 & 18 & 7/1/10 & classification \\
\bottomrule
\end{tabular}
\end{small}
\end{center}
\end{table}

\subsection{Correlation of Metrics}
\begin{figure}[!t]
    \centering        \includegraphics[width=0.95\columnwidth]{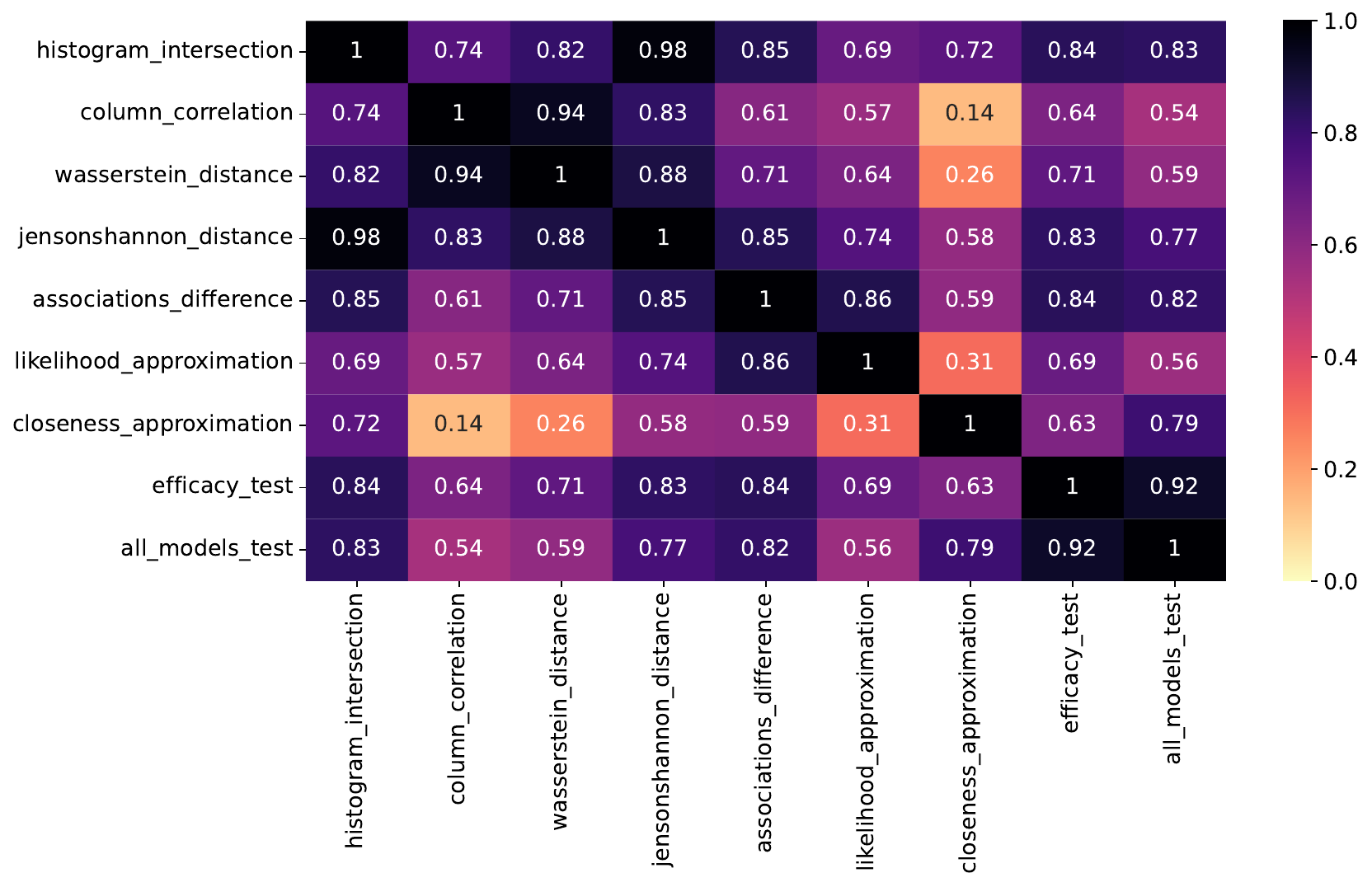}
    \vspace{-6pt}
    \caption{Pearson correlation coefficients (in absolute value) among different metrics across multiple experimental trials on all datasets.}
    \label{fig:metric_correlation_plot}
    \vspace{-10pt}
\end{figure}

We begin by a thorough correlation analysis of various metrics discussed in \sectionautorefname~\ref{metric}, as illustrated in \figureautorefname~\ref{fig:metric_correlation_plot}. Our observations revealed that metrics within each dimension generally exhibit a high degree of correlation. This is particularly evident in the case of the marginal-based metrics, as demonstrated by Pearson coefficients ranging between $0.74$ and $0.98$. Consequently, any of these metrics could adequately represent their respective dimension. For the sake of clarity and computational efficiency, we specifically chose to employ the \textbf{efficacy test (machine learning efficacy)}, \textbf{closeness approximation (distance to closest record)}, \textbf{associations difference}, and \textbf{histogram intersection} as representative metrics for summarizing each vital dimension discussed in \sectionautorefname~\ref{metric}.

\begin{figure*}
    \centering
    \includegraphics[width=\textwidth]{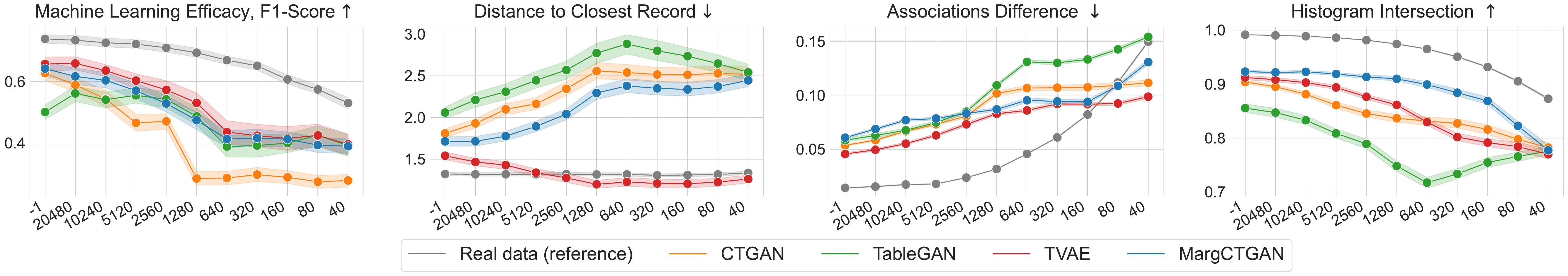}
    \caption{Averaged score across datasets.
    The \textbf{X-axis} represents the size of the training dataset, with ``-1'' indicating the full dataset size. \textbf{Real data (reference)} corresponds to the metrics directly measured on the real (train vs. test) data, serving as the reference (oracle score) for optimal performance.
    }
    \label{fig:averaged_metric_results}
\end{figure*}

\subsection{Methods Comparison}
\myparagraph{Downstream Utility}
The \textit{efficacy score} measures the utility of synthetic data in downstream tasks, as illustrated in the left-most plot of Figure~\ref{fig:averaged_metric_results}. In the best-case scenario (marked as ``-1'' in x-axis), the performances of $\ctgan$, $\tvae$, and $\mctgan$ are comparable. Performance generally degrades in low-sample settings, with the most significant drop around the size of $640$. This decline is particularly notable for $\ctgan$, which exhibits a relative error up to $57\%$. While $\tvae$ generally outperforms the other models across varying sample sizes, our $\mctgan$ performs robustly, demonstrating particular advantages in low-sample settings. Notably, $\mctgan$ consistently outperforms its backbone model, $\ctgan$, across all settings.

\myparagraph{Joint Fidelity \& Memorization}
The \textit{distance to the closest record} metric, depicted in the second subplot of Figure~\ref{fig:averaged_metric_results}, measures the alignment between the real and synthetic joint distribution and simultaneously illustrates the memorization effects of the generators.
Striking a balance is crucial as over-memorization might compromise privacy. $\tablegan$ consistently maintains the most substantial distance from the real data reference, aligning with its design objective of privacy preservation. Conversely, $\tvae$ displays the closest proximity, even exceeding the real reference, indicating a potential overfitting risk and privacy leakage. This may be attributed to its use of reconstruction loss in its training objective~\cite{chen2020gan}.
As the training size reduces, the distance between the synthetic and real data first increases then decreases, potentially signifying the generator's shift from generalization to memorization. While both $\ctgan$ and $\mctgan$ maintain a moderate distance from real data, our $\mctgan$ generally demonstrates a closer proximity to the reference, presenting an appropriate balance between alignment and privacy protection.

\myparagraph{Pairwise Correlation}
The \textit{association difference} metric (third subplot in \figureautorefname~\ref{fig:averaged_metric_results}) quantifies the disparity between the correlation matrices of the real and synthetic data. As expected, this disparity increases as the sample size decreases, a trend also seen in the real data reference. This could be attributed to data diversity, where different smaller subsets might not retain the same statistical characteristics while the sampling randomness is accounted for in our repeated experiments.
Among all models, $\tablegan$ exhibits the largest associations difference score, particularly in the low-sample regime, indicating challenges in capturing associations with limited training samples. Both $\mctgan$ and $\tvae$ display similar behavior, with our $\mctgan$ following the trend of real data reference more precisely, specifically in low-sample settings.

\begin{figure}[!t]
    \centering        \includegraphics[width=\columnwidth]{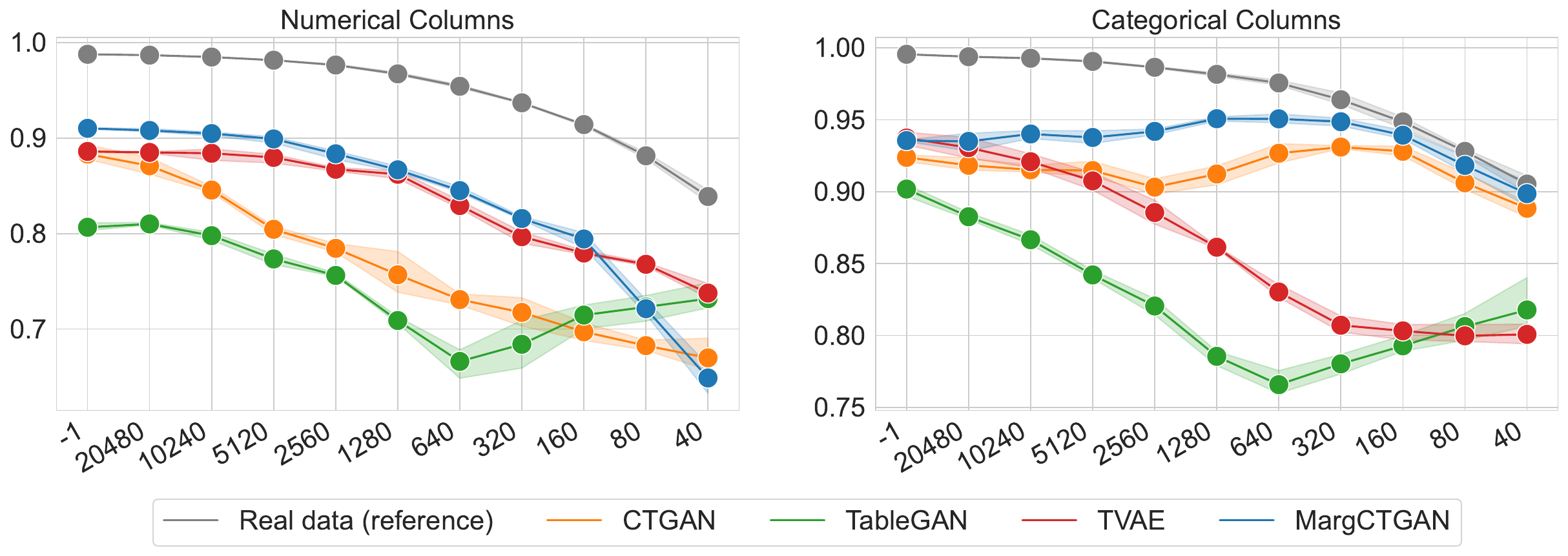}
    \caption{Histogram intersection score for numerical and categorical columns respectively, which is averaged across datasets.
    }
    \label{fig:averaged_histogram_intersection_numerical_categorical_e300}
\end{figure}

\myparagraph{Marginal Matching}
The \textit{histogram intersection} metric, depicted in the right-most subfigure in \figureautorefname~\ref{fig:averaged_metric_results}, assessing the overlap of real and synthetic marginal distributions. 
Our moment matching objective within $\mctgan$ explicitly encourages such coverage of low-level statistics, leading to consistent superior performance of $\mctgan$ across various settings. A more detailed analysis, presented in \figureautorefname~\ref{fig:averaged_histogram_intersection_numerical_categorical_e300}, reveals performance differences across numerical and categorical columns. Here, $\tvae$ demonstrates good performance with numerical attributes but exhibits limitations in handling categorical ones, whereas $\ctgan$ excels in handling categorical columns, possibly due to its \textit{training-by-sampling} approach, but falls short with the numerical ones.
Notably, $\mctgan$ balances both aspects, outperforming $\ctgan$ in numerical columns while matching its performance in categorical ones. 
Moreover, while most models show decreased performance in low-resource settings, $\tablegan$ exhibits improvement, potentially due to its similar moment matching approach to ours,
thereby further validating our design choice.

\section{Conclusion}
\label{conclusion}
In conclusion, our comprehensive evaluation of state-of-the-art synthetic tabular data generators provided valuable insights, particularly regarding their performance in low-resource settings. By introducing the $\mctgan$ model as an improvement over $\ctgan$, we addressed some of the limitations observed. Our results demonstrated that $\mctgan$ outperformed $\ctgan$ across multiple metrics, exhibiting stronger utility and marginal-based performance. These findings highlight the effectiveness of $\mctgan$ in enhancing both the utility and marginal score of the generated synthetic data. Our code and setup is published to ensure reproducibility and increase impact\footnote{https://github.com/tejuafonja/margctgan}.

\section*{Acknowledgement}
\label{acknowledgement}
This work was partially funded by ELSA – European Lighthouse on Secure and Safe AI funded by the European Union under grant agreement No. 101070617. Additionally, this work is supported by the Helmholtz Association within the project ``Protecting Genetic Data with Synthetic Cohorts from Deep Generative Models (PRO-GENE-GEN)'' (ZT-I-PF-5-23) and Bundesministeriums für Bildung und Forschung ``PriSyn'' (16KISAO29K). Views and opinions expressed are however those of the authors only and do not necessarily reflect those of the European Union or European Commission. Neither the European Union nor the European Commission can be held responsible for them. Moreover, Dingfan Chen was partially supported by Qualcomm Innovation Fellowship Europe.

\bibliography{main.bib}
\bibliographystyle{icml2023}

\newpage
\appendix
\onecolumn
\section{Implementation Details}
\label{implementation_details}
All the code was implemented in the Python and all experiments are conducted on a single Titan RTX GPU. 

\vspace{-4pt}
\subsection{Metrics}
\paragraph{Application Utility.}
For the \textit{machine learning efficacy} and \textit{all models test} metrics, we implemented them using the SDMetrics package~\footnote{\url{https://docs.sdv.dev/sdmetrics/metrics/metrics-in-beta/ml-efficacy-single-table/binary-classification}}. We trained logistic regression, decision tree classifier, and multilayer perceptron models for the classification tasks, and linear regression, decision tree regressor, and multilayer perceptron models for the regression tasks. We kept the default hyperparameter settings. For the classification models, we standardized the numerical columns during training, while for the regression models, no normalization scheme was applied as it yielded more stable results without it. The categorical columns for both regression and classification tasks were one-hot encoded. Each classification model was evaluated using the F1\text{-score} while each regression model is evaluated with $R^2\text{-score}$. The $R^2\text{-score}$ has a range of $[-\infty, 1]$, so we normalized it to $[0, 1]$ using the normalization scheme from the SDMetrics package.

\vspace{-4pt}
\paragraph{Joint Fidelity.}
We applied a min-max normalization scheme to all numerical columns to constrain their value range between 0 and 1. Additionally, we one-hot encoded the categorical columns. We used the scikit-learn nearest-neighbor implementation~\footnote{\url{https://scikit-learn.org/stable/modules/generated/sklearn.neighbors.KNeighborsClassifier.html}} and chose the number of nearest neighbors in the range of $[1,2,3,\ldots,9]$ while using the Euclidean distance. For the \textit{likelihood approximation}, we calculated the distance between each of 5000 random test samples to its closest synthetic sample and report the average over real test samples. For the \textit{distance to closest record} metric, we compute the distance of each sample in the synthetic set to its nearest neighbor in a set of 5000 random test samples and report the average over the synthetic samples. We use the default hyperparameters from the package.

\vspace{-4pt}
\paragraph{Column-Pair Fidelity.}
The \textit{associations difference} metric was inspired by the ``plot correlation difference'' function from the tabular-evaluator package~\footnote{\url{https://github.com/Baukebrenninkmeijer/table-evaluator}} and implemented using the dython package~\footnote{\url{http://shakedzy.xyz/dython/modules/nominal/}}. We used Pearson correlation coefficient for numerical columns, Cramer's V for categorical columns, and the Correlation Ratio for numerical-categorical columns. The range of Cramer's V and Correlation Ratio is between 0 and 1, while Pearson correlation coefficient ranges from -1 to 1. We calculated the absolute difference between the association matrices of the synthetic data and the real data, reporting the mean absolute difference.

\vspace{-4pt}
\paragraph{Marginal Fidelity.}
We applied a min-max normalization scheme to all numerical columns to constrain their value range between 0 and 1, and label-encoded the categorical columns. The marginal metrics were applied to each column in the dataset. To speed up the calculation, we parallelized the process with joblib~\footnote{\url{https://joblib.readthedocs.io/en/latest/}}. The \textit{histogram intersection}, \textit{Wasserstein distance}, and \textit{Jenson-Shannon distance} used the same binning strategy for the numerical columns. We assumed a uniform grid between 0 and 1 and computed the bin width $\bigtriangleup_i$ for the real data (train and test) using a pre-defined bin sizes $\in \{25,50,100\}$. We also used the real data (train and test) to determine the number of categories for each categorical column. The categorical columns has $\bigtriangleup_i=1$. The normalized count for each bin was calculated for the synthetic and real data. We used the SciPy package~\footnote{\url{https://docs.scipy.org/}} to calculate \textit{Wasserstein distance} by passing the grids as empirical values and the normalized bin values as weights, and to calculate \textit{Jenson-Shannon distance} with the base=2 setting. We used the dython package to compute \textit{column correlation} metric. As for the \textit{histogram intersection} metric, we implemented it ourselves. Unless otherwise specified, the hyperparameter settings were taken from the packages.

\vspace{-4pt}
\subsection{Generative Models}
We used the open-source implementation provided by the authors for $\ctgan$ and $\tvae$~\cite{ctgan} and the SDGym package\footnote{\url{https://github.com/sdv-dev/SDGym/tree/41ffcc1982e2c5081271ce0e138792550c2a3480/sdgym/synthesizers}} for $\tablegan$. For the $\mctgan$, we modified the $\ctgan$ model and follow the default hyperparameter setting. We use the same number of training epochs and batch size  for all models. The data generators were trained on each subset of the datasets, and after training, the synthetic datasets were sampled. To ensure robustness, we repeated this process three times using different random seeds.

To evaluate the performance of the generators, we sampled five synthetic datasets with 20,000 samples from each data synthesizer. We fitted utility and joint-based metrics on the synthetic datasets and then evaluated the fitted model against the real test dataset. For the column-pair-based and marginal-based metrics, we first computed statistics for the synthetic datasets and real test dataset separately and then compared the statistics for evaluation. For the real data reference (always shown as gray lines in the figures), we run the same evaluation process as above but replace the synthetic set by the real training set.

\section{Datasets}
\label{dataset_info}

\myparagraph{Adult~\footnote{\url{https://archive.ics.uci.edu/ml/datasets/adult}}} The UCI Adult dataset~\cite{adult} is from the 1994 U.S. Census and contains 14 attributes with a total sample size of 48,000. The associated task is a binary classification to determine if an individual earns more than 50,000 dollars a year. 

\myparagraph{Census~\footnote{\url{https://archive.ics.uci.edu/ml/datasets/US+Census+Data+\%281990\%29}}} The UCI Census-income dataset~\cite{census} is a weighted dataset of the 1994-5 U.S. Census conducted by the U.S. Census Bureau. It includes over 299,000 samples and 40 attributes with an associated binary classification task of determining whether a person earns more than 50,000 dollars a year.

\myparagraph{News~\footnote{\url{https://archive.ics.uci.edu/ml/datasets/online+news+popularity}}} The UCI Online News Popularity dataset~\cite{news} summarizes a heterogeneous set of features about articles published by Mashable over a two-year period. It includes over 39,000 instances with 58 attributes whose goal is to predict the number of shares on social networks (we dropped the non-predictive attributes). It is a regression task, but it can be transformed into a classification task. As executed in \citet{news}, we assume a binary classification task, where an article is considered ``popular" if the number of shares is higher than a fixed decision threshold $t$, else it is ``unpopular''. $t=1400$.

\myparagraph{Texas~\footnote{\url{https://github.com/spring-epfl/synthetic_data_release/blob/master/data/texas.csv}}} The Texas Hospital Discharge dataset is a large public use data file provided by the Texas Department of State Health Services~\footnote{\url{https://www.dshs.texas.gov/thcic/}}. We used the preprocessed version from \citet{syntheticgroundhog}, which consists of 100,000 records uniformly selected from a pre-processed file containing patient data from 2013. We retain 18 attributes and assume a binary classification task by predicting only minor and major mortality risk. The final size of the dataset was therefore reduced to 75,159.

\section{Extended Results}
\subsection{ Insights into Performance and Behavior of Histogram Intersection Metric}
\figureautorefname~\ref{fig:averaged_histogram_intersection_numerical_categorical_e300} highlighted that $\tvae$ excels in datasets with numerical columns but struggles with categorical columns, resulting in subpar marginal performance due to its inability to accurately reproduce different categories. Surprisingly, despite this limitation, the synthetic datasets generated by $\tvae$ showed high utility in downstream tasks, suggesting that a good marginal distribution is not always a prerequisite for usefulness. On the other hand, $\ctgan$ demonstrated superior performance in capturing associations within categorical columns, showcasing the effectiveness of its \textit{training-by-sampling} approach. Notably, $\mctgan$ achieved similar performance to $\ctgan$ in categorical columns while outperforming it in numerical columns, aligning more closely with $\tvae$ in this aspect.

\begin{figure}[!h]
    \centering    
    \subfigure[Adult]{
    \includegraphics[width=0.45\textwidth]{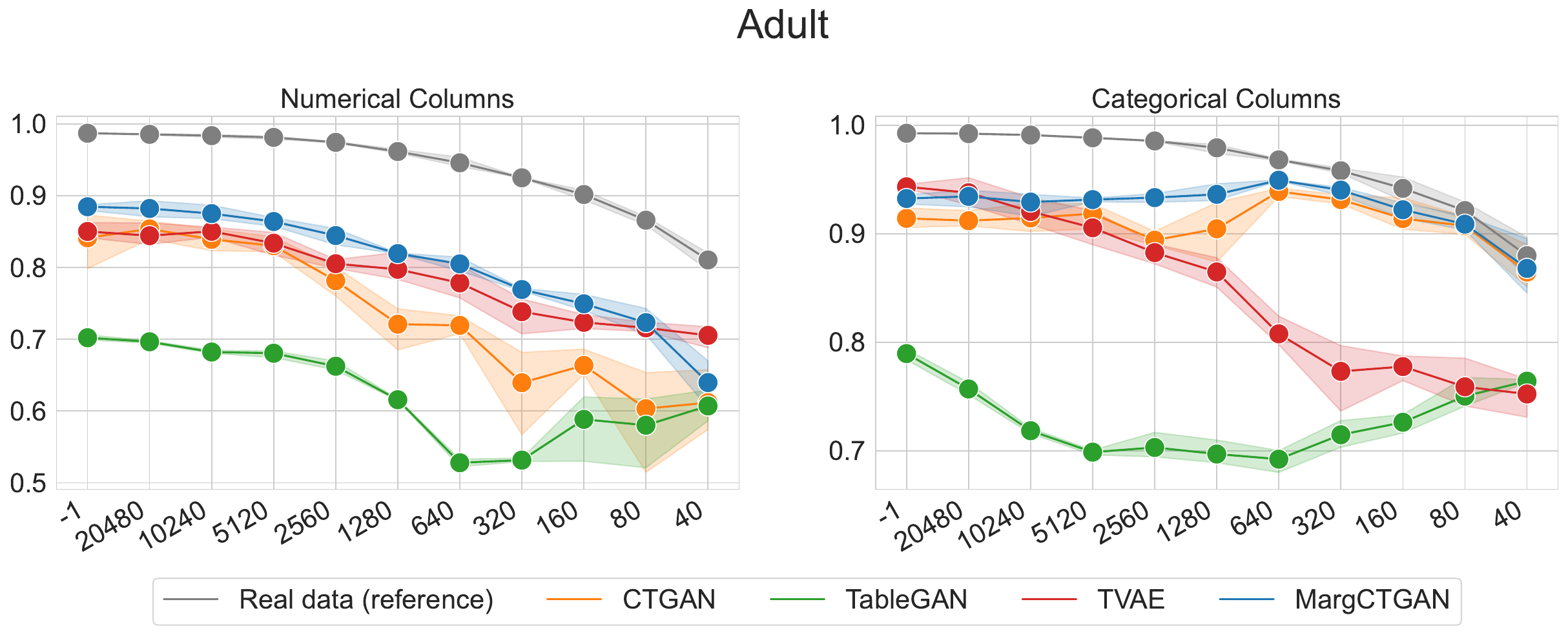}
    \label{fig:adult_histogram_intersection_numerical_categorical_e300}}
     \subfigure[Census]{
     \includegraphics[width=0.45\textwidth]{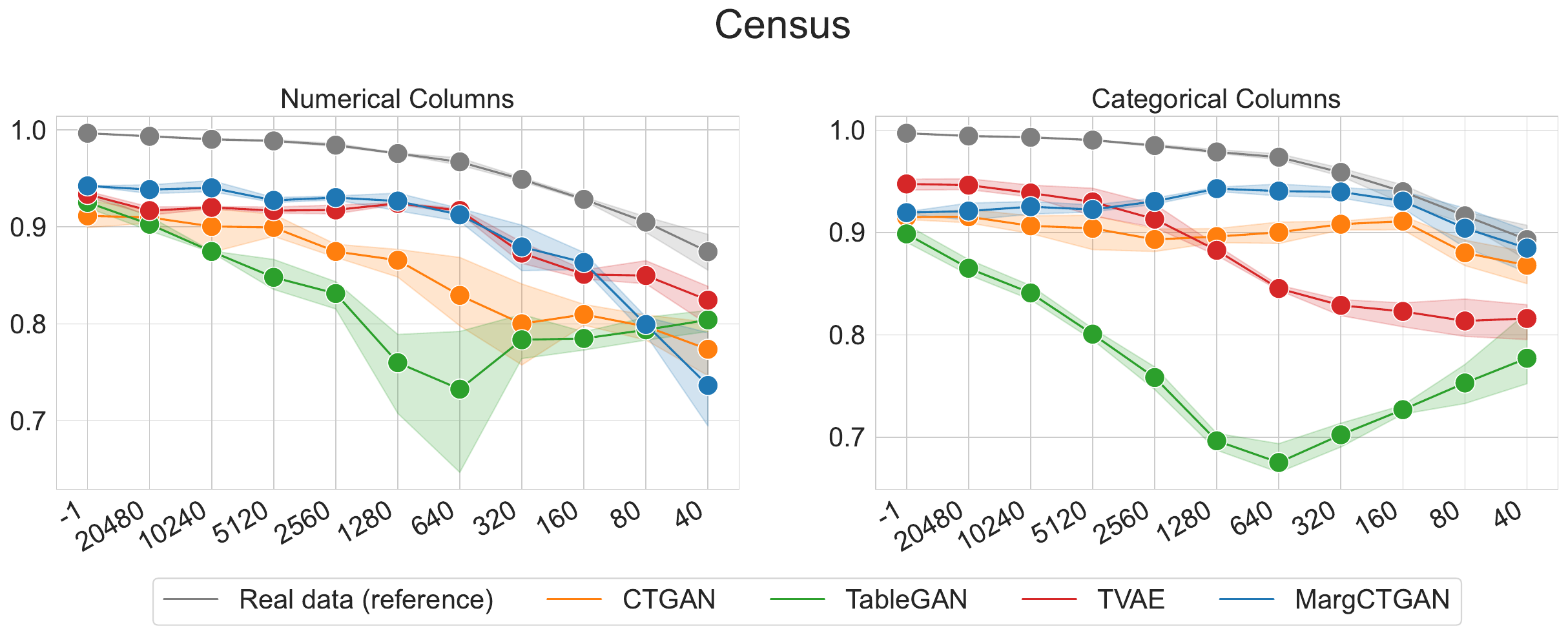}
    \label{fig:census_histogram_intersection_numerical_categorical_e300}}
    \subfigure[News]{
    \includegraphics[width=0.45\textwidth]{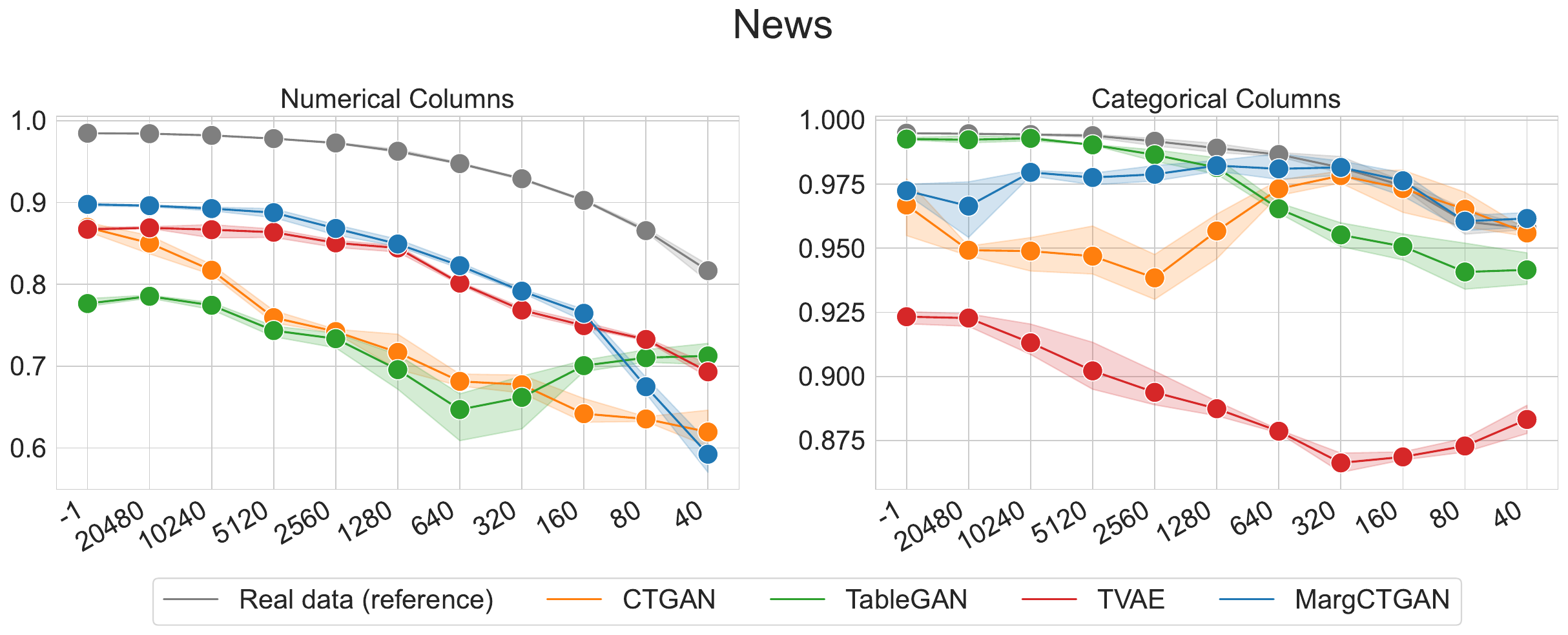}
    \label{fig:news_histogram_intersection_numerical_categorical_e300}}
    \subfigure[Texas]{
     \includegraphics[width=0.45\textwidth]{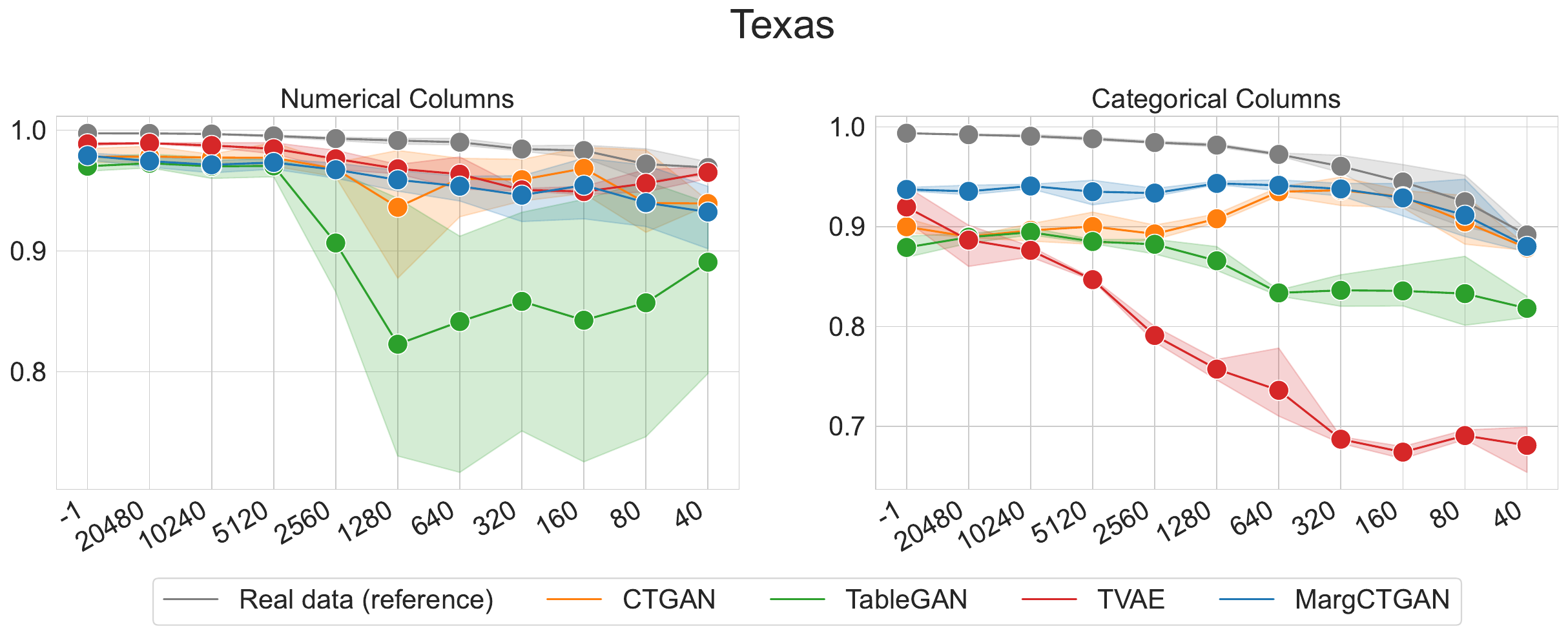}
    \label{fig:texas_histogram_intersection_numerical_categorical_e300}}
    \caption{Histogram intersection score on  Adult~\ref{fig:adult_histogram_intersection_numerical_categorical_e300}, Census~\ref{fig:census_histogram_intersection_numerical_categorical_e300},   News~\ref{fig:news_histogram_intersection_numerical_categorical_e300}, and Texas~\ref{fig:texas_histogram_intersection_numerical_categorical_e300} dataset, respectively.}
\end{figure}

It is important to note that all models experienced degraded performance in low-resource settings. However, interestingly, $\tablegan$ exhibited improved performance in such scenarios. This improvement may be attributed to its \textit{information loss}, which share similar idea of our moment matching objective. 
Figures~\ref{fig:adult_histogram_intersection_numerical_categorical_e300}, ~\ref{fig:census_histogram_intersection_numerical_categorical_e300},  ~\ref{fig:news_histogram_intersection_numerical_categorical_e300}, ~\ref{fig:texas_histogram_intersection_numerical_categorical_e300} further shows the breakdown across the different datasets.

\subsection{Relative Error}
We additionally provide the \textbf{relative error} w.r.t. \textit{real data (reference)} for each metric (averaged across datasets) in \tableautorefname~\ref{tab:relative_error_mle}, \ref{tab:relative_error_closeness_approximation}, \ref{tab:relative_error_associations_difference}, and \ref{tab:relative_error_histogram_intersection}.

\begin{table}[!h]
    \centering
    \small
    \resizebox{0.9\columnwidth}{!}{
    \begin{tabular}{lrrrrrrrrrrr}
    \toprule
    subset &     -1 &  20480 &  10240 &   5120 &   2560 &   1280 &    640 &    320 &    160 &     80 &     40 \\
    \midrule
    $\tablegan$      &  32.24 &  23.61 &  25.55 &  23.13 &  23.65 &  29.90 &  42.05 &  39.89 &  34.19 &  26.03 &  26.21 \\
    $\tvae$          &  11.03 &  10.29 &  12.49 &  16.50 &  19.30 &  23.54 &  34.96 &  34.99 &  31.65 &  26.08 &  25.40 \\
    $\ctgan$         &  15.07 &  19.81 &  25.25 &  35.55 &  33.75 &  59.07 &  57.35 &  54.45 &  52.57 &  52.38 &  47.66 \\
    $\mctgan$     &  13.14 &  16.04 &  16.89 &  21.05 &  25.56 &  31.72 &  38.36 &  36.14 &  32.13 &  31.55 &  26.57 \\
    \bottomrule
    \end{tabular}}
    \caption{Averaged relative error of \textit{machine learning efficacy}. The lower the better.}
    \label{tab:relative_error_mle}
\end{table}

\begin{table}[!h]
    \centering
    \small
    \resizebox{0.9\columnwidth}{!}{
    \begin{tabular}{lrrrrrrrrrrr}
    \toprule
    subset &     -1 &  20480 &  10240 &   5120 &   2560 &    1280 &     640 &     320 &     160 &      80 &     40 \\
    \midrule
    $\tablegan$      & -55.76 & -67.57 & -74.93 & -85.25 & -94.39 & -110.51 & -118.69 & -114.20 & -108.63 & -100.92 & -90.18 \\
    $\tvae$          & -16.59 & -11.23 &  -8.28 &  -1.41 &   3.44 &    8.92 &    6.89 &    7.49 &    7.95 &    6.99 &   5.49 \\
    $\ctgan$         & -36.92 & -46.24 & -58.97 & -63.89 & -77.39 &  -94.15 &  -92.71 &  -92.43 &  -91.71 &  -92.02 & -88.41 \\
    $\mctgan$     & -29.65 & -29.96 & -34.60 & -43.51 & -54.38 &  -74.16 &  -80.52 &  -79.75 &  -78.25 &  -80.00 & -82.99 \\
    \bottomrule
    \end{tabular}}
    \captionsetup{width=0.8\linewidth}
    \caption{Averaged relative error of \textit{closeness approximation}. Generally, a lower value is considered better. However, an excessively low value may suggest memorization.}
\label{tab:relative_error_closeness_approximation}
\end{table}

\begin{table}[!h]
    \centering
    \small
    \resizebox{0.9\columnwidth}{!}{
    \begin{tabular}{lrrrrrrrrrrr}
    \toprule
    subset &      -1 &   20480 &   10240 &    5120 &    2560 &    1280 &     640 &     320 &    160 &     80 &     40 \\
    \midrule
    $\tablegan$      & -312.05 & -308.51 & -291.45 & -319.78 & -258.71 & -245.16 & -186.87 & -113.52 & -62.23 & -27.60 &  -2.90 \\
    $\tvae$          & -221.95 & -222.38 & -219.46 & -253.43 & -209.10 & -161.99 &  -88.33 &  -51.00 & -11.69 &  17.34 &  34.10 \\
    $\ctgan$         & -278.87 & -280.84 & -286.25 & -311.11 & -241.34 & -221.51 & -133.71 &  -75.76 & -30.71 &   2.29 &  25.52 \\
    $\mctgan$     & -328.64 & -347.45 & -345.54 & -340.73 & -251.48 & -174.78 & -108.87 &  -54.71 & -14.36 &   2.76 &  12.69 \\
    \bottomrule
    \end{tabular}}
    \caption{Averaged relative error of \textit{associations difference}. The lower the better.}
    \label{tab:relative_error_associations_difference}
\end{table}

\begin{table}[!h]
    \centering
    \small
    \resizebox{0.9\columnwidth}{!}{
    \begin{tabular}{lrrrrrrrrrrr}
    \toprule
    subset &     -1 &  20480 &  10240 &   5120 &   2560 &   1280 &    640 &    320 &    160 &     80 &     40 \\
    \midrule
    $\tablegan$      &  13.72 &  14.44 &  15.74 &  17.98 &  19.59 &  23.21 &  25.66 &  22.84 &  18.98 &  15.40 &  11.10 \\
    $\tvae$          &   7.99 &   8.27 &   8.68 &   9.34 &  10.69 &  11.58 &  14.03 &  15.64 &  15.03 &  13.38 &  11.78 \\
    $\ctgan$         &   8.83 &   9.60 &  10.87 &  12.67 &  13.85 &  14.13 &  13.83 &  12.98 &  12.42 &  11.89 &  10.38 \\
    $\mctgan$     &   6.90 &   6.92 &   6.67 &   6.82 &   6.93 &   6.63 &   6.82 &   7.01 &   6.74 &   9.13 &  10.94 \\
    \bottomrule
    \end{tabular}}
    \caption{Averaged relative error of \textit{histogram intersection}. The lower the better.}
    \label{tab:relative_error_histogram_intersection}
\end{table}

\newpage
\vspace{-8pt}
\subsection{Metrics Results for Different Datasets}
\begin{figure*}[!h]
    \centering
    \subfigure[Adult]{
    \includegraphics[width=\textwidth]{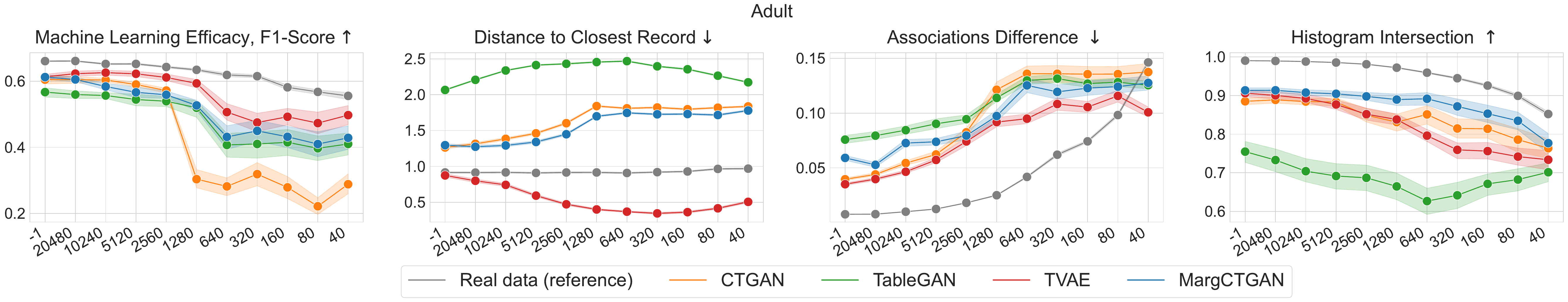}
    \label{fig:adult_combined_e300}}
    \subfigure[Census]{ \includegraphics[width=\textwidth]{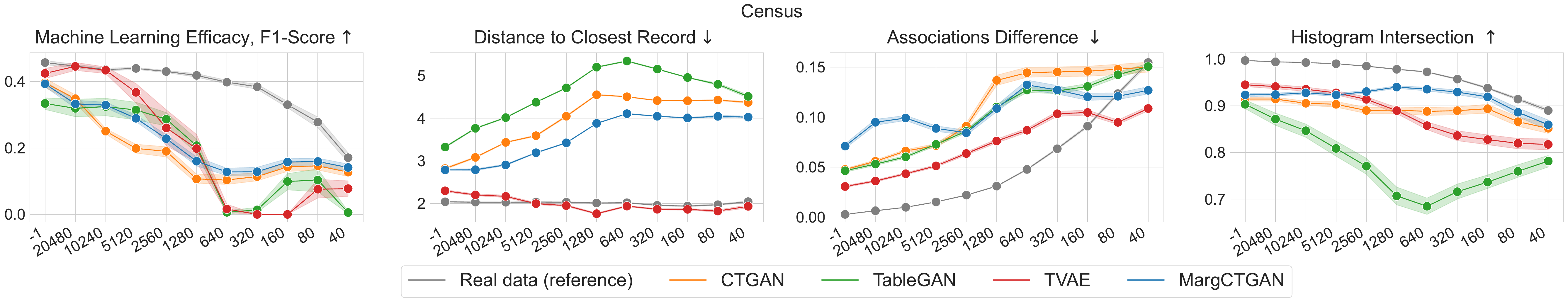}
    \label{fig:census_combined_e300}}
 \subfigure[News]{\includegraphics[width=\textwidth]{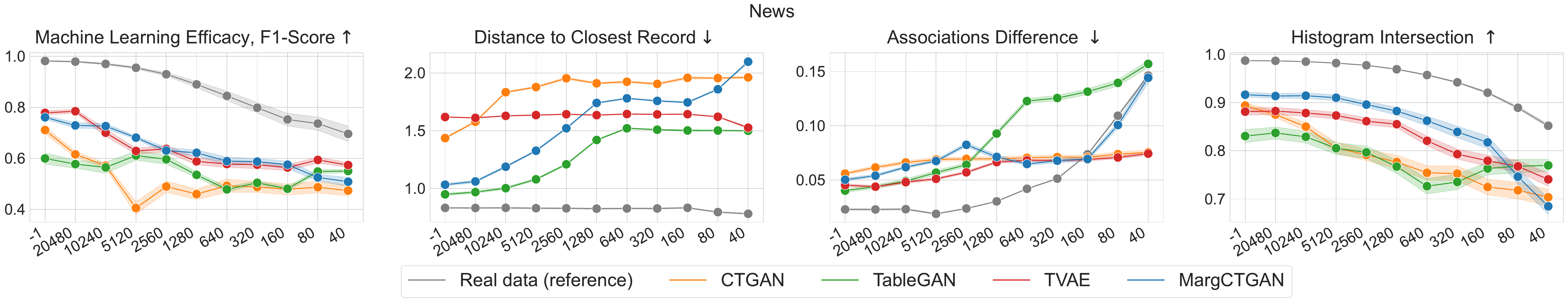}
 \label{fig:news_combined_e300}
 }
 \subfigure[Texas]{\includegraphics[width=\textwidth]{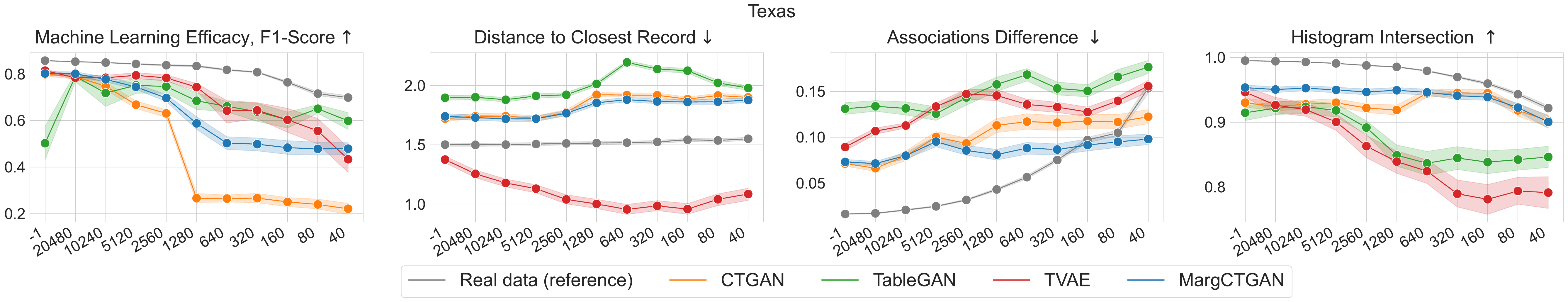}
    \label{fig:texas_combined_e300}}
 \caption{Metric evaluation results on each dataset, which is supplementary to \figureautorefname~\ref{fig:averaged_metric_results} in the main paper.
 }
\end{figure*}

\figureautorefname~\ref{fig:adult_combined_e300}, \ref{fig:census_combined_e300}, \ref{fig:news_combined_e300}, and \ref{fig:texas_combined_e300} show the 
results of metric evaluation on each dataset respectively.

\vspace{-8pt}

\newpage
\section{Ablation}
\subsection{Moment Matching in Raw Data Space}
\label{performance_in_raw_space}
\begin{figure*}[!h]
    \centering
    \includegraphics[width=\textwidth]{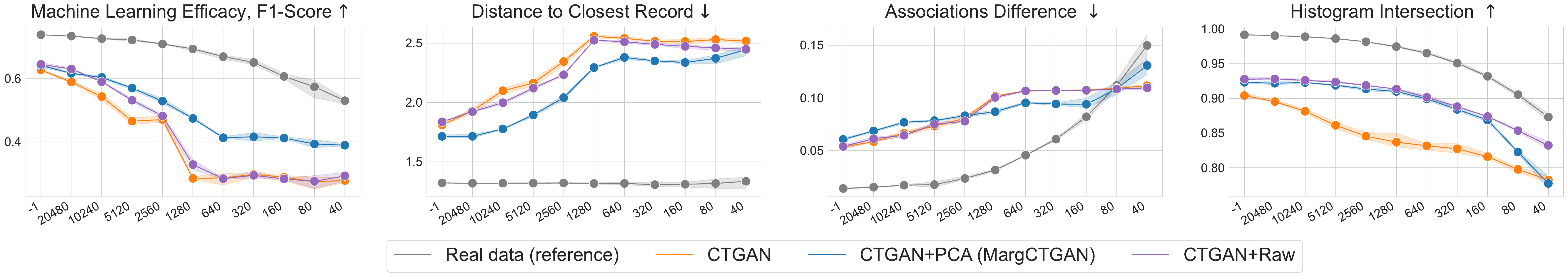}
    \caption{Comparison between $\ctgan$ trained with PCA loss objective ($\mctgan$) and $\ctgan$ trained with raw moment matching loss objective.}
    \label{fig:ablation}
\end{figure*}

We conducted an additional ablation study to investigate the effect of the moment matching technique with and without applying PCA in $\mctgan$. As shown in \figureautorefname~\ref{fig:ablation}, while both moment matching without PCA ($\texttt{CTGAN+Raw}$) and with PCA ($\mctgan$) performs generally better than the baseline $\ctgan$, the PCA adopted in our $\mctgan$ does provide additional notable improvement consistently across different metrics  considered in our study.

\subsection{Training for Longer Epoch}
\label{training_longer}
\begin{figure*}[!h]
    \centering
    \subfigure[Adult, Epoch=300]{
    \includegraphics[width=\textwidth]{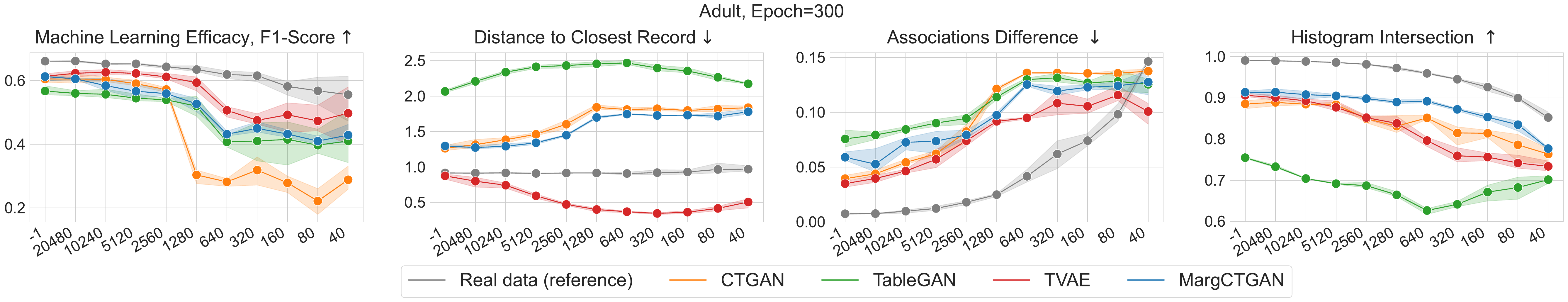}
    \label{fig:adult_combined_ablation_e300}}
    \subfigure[Adult, Epoch=1000]{ \includegraphics[width=\textwidth]{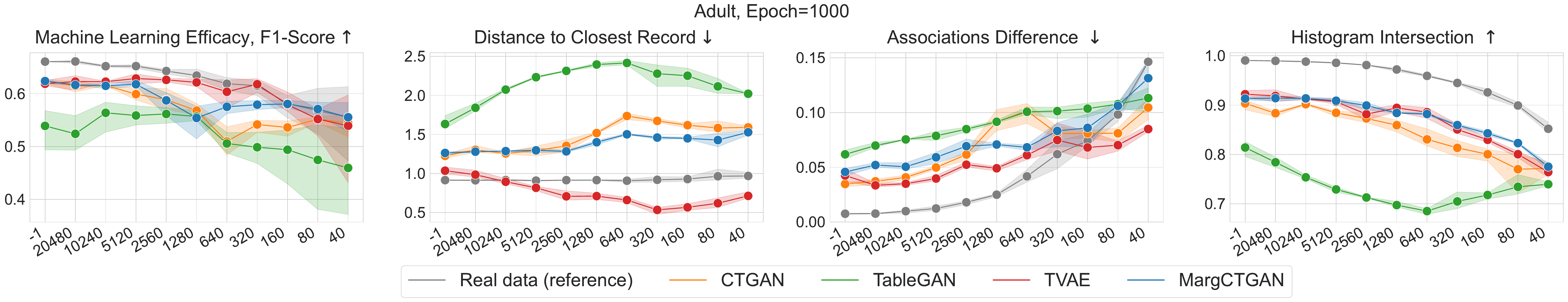}
    \label{fig:adult_combined_ablation_e1000}}
\caption{Comparison showing performance of Adult dataset trained for 300 (top) and 1000 (bottom) epochs.}
\end{figure*}

Previous work \cite{karlsson2020synthesis} has shown that the performance of $\ctgan$ generally improves with further training, and we validate this finding in our experiments. Nevertheless, we limited the training epochs to 300 to align with the specified hyperparameters by the authors, considering the common usage of these models as off-the-shelf solutions without extensive tuning. Despite this constraint, our $\mctgan$ model consistently outperforms $\ctgan$ as shown in Figure ~\ref{fig:adult_combined_ablation_e300}, and  ~\ref{fig:adult_combined_ablation_e1000}. This highlights the enhanced performance of $\mctgan$ in comparison to $\ctgan$ in synthetic data generation tasks.

\vspace{-8pt}

\section{Limitations}
The evaluation of \textit{associations difference} uncovered insightful observation regarding the performance of $\mctgan$ in low-resource settings. As the sample size decreased to 40, the performance of $\mctgan$ exhibited degradation, while $\ctgan$ demonstrated better results for this metric. A similar trend was observed for the marginal-based metric. This behavior can be attributed to the inherent characteristics of the PCA-moment matching loss method employed by $\mctgan$. The effective number of PCA components used in the transformation is determined by the minimum value between the size of the data and the size of the feature. In other words, the method can accurately capture features up to the limit set by the smaller value of these two factors. Consequently, in scenarios where the sample size is smaller than the number of features, the PCA-moment matching approach becomes less competitive (due to the rank-deficiency during the matrix computation). This limitation arises from the method's inability to effectively match and reproduce the characteristics of all features due to the constrained number of components available. Consequently, the performance of $\mctgan$ deteriorates in terms of both the associations difference metric and the marginal-based metric. Supporting evidence for this limitation can be found in the experiment conducted with raw features (see Appendix~\ref{performance_in_raw_space}). It is crucial to consider this constraint when selecting an appropriate model for synthetic data generation. In scenarios where the sample size is significantly smaller than the number of features, alternative models such as $\texttt{CTGAN+Raw}$ may offer better performance in capturing associations and reproducing marginal distributions. Understanding the strengths and weaknesses of different models under varying resource constraints empowers researchers and practitioners to make informed decisions and select the most suitable synthetic data generation approach for their specific requirements.


\end{document}